\documentclass[11pt]{article}

\usepackage{array}
\usepackage{subcaption}
\usepackage{multirow}
\usepackage{booktabs}
\usepackage{amsmath}
\usepackage{amssymb}
\usepackage{natbib}
\usepackage{graphicx}
\usepackage{authblk}
\usepackage[margin=1in]{geometry}
\usepackage{hyperref}

\title{Joint distribution of upstream runoff governs downstream river-discharge prediction uncertainty in distributed ML models}

\author[1,2,3]{Karan Ruparell\thanks{Corresponding author: k.ruparell2@pgr.reading.ac.uk}}
\author[3,6]{Tristan Hascoet}
\author[3,6]{Takemasa Miyoshi}
\author[1,4]{Kieran M. R. Hunt}
\author[1,5]{Hannah L. Cloke}
\author[2]{Christel Prudhomme}
\author[2]{Florian Pappenberger}

\affil[1]{University of Reading, Department of Meteorology, Reading, UK}
\affil[2]{European Centre for Medium-Range Weather Forecasts (ECMWF), Reading, UK}
\affil[3]{RIKEN Center for Computational Sciences, Kobe, Japan}
\affil[4]{National Centre for Atmospheric Science (NCAS), Reading, UK}
\affil[5]{University of Reading, Department of Geography and Environmental Science, Reading, UK}
\affil[6]{RIKEN Center for Interdisciplinary Theoretical and Mathematical Sciences, Kobe, Japan}

\date{}

\begin{document}
\maketitle

\begin{abstract}
Uncertainty quantification of hydrological predictions is necessary to inform operational decisions. Recent generative machine-learning methods have advanced probabilistic streamflow prediction, but have remained confined to lumped models that predict a basin outlet directly. At the same time, deterministic LSTM runoff models are increasingly applied at grid or catchment scale and routed through river networks to produce spatially continuous, physically consistent discharge fields. This technical note argues that moving probabilistic prediction from lumped to distributed models introduces a specific new requirement: the joint distribution of upstream runoff generation must be sampled jointly. In lumped inference, the model predicts the outlet distribution directly and can modulate spread from basin attributes. In distributed inference, downstream discharge is obtained by routing many upstream runoff predictions, so independent local sampling averages uncertainty away. Using Japan as a case study, we train two probabilistic basin-scale runoff LSTMs and route their runoff through a Hayami routing scheme. Randomly matching upstream ensemble members produces severely under-dispersed downstream ensembles, whereas a simple quantile matching strategy restores much of the spread of the direct basin-scale reference. The shift from lumped to distributed probabilistic hydrology therefore requires explicit attention to the spatial joint structure of runoff uncertainty.

\end{abstract}

\section{Plain Language Summary}
Predicting river flow is not just about forecasting a single, most-likely value. It also requires providing a realistic range of possible outcomes so water managers can make informed decisions. Recently, artificial intelligence (AI) has become very good at predicting these probability ranges for specific river locations. However, as hydrologists expand these AI models to predict flow across entire, interconnected river networks, a new problem arises. 

A river's final flow can often be approximated as the sum of water coming from many smaller upstream areas. If an AI model assumes that the uncertainty in each of these upstream areas is independent, those errors cancel each other out as the water flows downstream. This results in a final forecast that is overly confident and too narrow. Using rivers in Japan as a case study, we demonstrate this "averaging out" problem and test a solution. We show that to get realistic, dispersed forecasts at the river outlet, the AI must link—or "jointly sample"—the uncertainties across the upstream areas. To build trustworthy AI models for whole river networks, we must correctly capture how runoff uncertainty is spatially connected.

\section{Introduction}
Ensemble forecasts have long been recognised as vital for hydrological forecasting \citep{cloke2009ensemble, troin2021generating}, motivating initiatives such as the Hydrological Ensemble Prediction Experiment (HEPEX)
\citep{schaake2007hepex}. For operational river-discharge prediction, a well-calibrated estimate of uncertainty necessary for decision-makers to weigh the risk of flooding, allocate reservoir storage, or issue evacuation orders.

Machine-learning (ML) models have rapidly become competitive for streamflow prediction \citep{kratzert2018rainfall, hunt2022using, feng2020enhancing}, and recent work has pushed them from deterministic forecasts towards full probabilistic prediction: regressing quantiles of forecast probability \citep{ruparell2025hydra}, modelling per-leadtime distributions of river flow \citep{klotz2022uncertainty}, and generating temporally consistent ensemble trajectories \citep{ruparell2026ai, kraft2026modeling}. These methods capture uncertainty from hydrological states, catchment attributes, and latent model bias without requiring a separate ensemble of atmospheric drivers. They have, however, been developed mostly in the \emph{lumped} setting, where a model predicts the discharge distribution at a basin outlet directly.


At the same time, deterministic ML hydrology is moving towards spatially distributed prediction. Large-sample basin datasets such as Caravan support training LSTMs on many basins \citep{kratzert2023caravan}; distributed systems such as GRADES-hydroDL \citep{yang2025global} then apply deterministic LSTM runoff models over grids or local hydrological units and route the generated runoff through explicit river networks \citep{feng2020enhancing,bindas2024improving,hascoet2026differentiable, mosaffa2025gnn}. 

Unlike a collection of independent lumped models, a routed system conserves water balance across the network, ensures that flow accumulates from headwaters to outlet as it must, and allows us to make predictions at ungauged locations throughout the channel. A lumped ensemble at monitored outlets has no mechanism preventing it from predicting higher discharge at an upstream gauge than at a downstream one, and so can produce physically impossible results that routing eliminates by construction. For these reasons, distributed prediction are valuable even if they reduce per-gauge NSE or CRPS. 

In fact, a lumped model trained directly at a basin outlet can implicitly learn the integrated hydrological response, so can often achieve higher per-gauge skill than a distributed model whose local runoff estimates are assembled and routed. Because of this inherent advantage, we use the lumped model not as a competitor, but as a performance upper-bound for per-gauge uncertainty quantification. This allows us to isolate and measure how much value our spatial sampling strategies add to the distributed model relative to a theoretical ceiling.

Simply replacing the deterministic runoff model with a probabilistic one is not sufficient to produce well-calibrated distributed ensemble forecasts. In a lumped model, the outlet distribution is learned as a function of basin-scale inputs and attributes, and the model can implicitly modulate its spread to reflect basin size and hydrological regime. In a distributed model, downstream discharge is the routed sum of runoff from many upstream catchments. This means that the spread of the resulting ensemble depends not only on each catchment's marginal runoff distribution but on how ensemble members are matched across catchments. It is determined by the joint distribution of the upstream runoff samples: which upstream catchments are high, low, or near their mean in the same ensemble member. This joint structure is implicit in lumped inference but must be specified explicitly in distributed inference.

\begin{figure}
    \centering
    \includegraphics[width=0.9\linewidth]{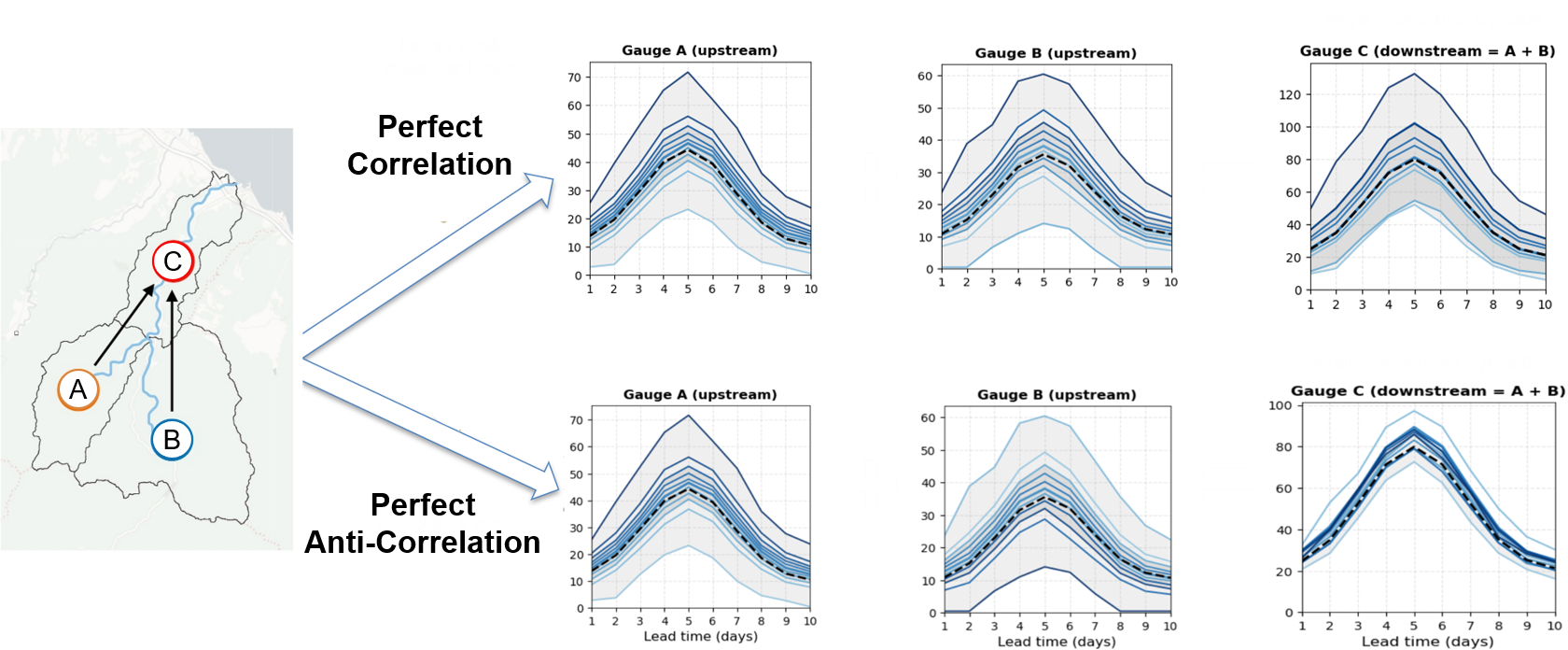}
    \caption{Why the joint structure of runoff uncertainty matters. A simple
    network routes runoff from two upstream gauges (A and B) into a downstream
    gauge (C $=$ A $+$ B). Each upstream gauge carries the same marginal ensemble
    spread, but the downstream spread depends on how the two are correlated across
    ensemble members. Under a large-scale driver (top), errors at A and B are
    positively correlated and compound under routing into a wide, well-calibrated
    downstream spread. Under a small-scale, localised driver (bottom), errors are
    uncorrelated or anti-correlated and cancel under routing, producing a narrow,
    under-dispersed downstream ensemble.}
    \label{fig:motivation}
\end{figure}

Consider the network of Fig.~\ref{fig:motivation}, where catchments A and B each contribute runoff to a downstream gauge C, and both upstream ensembles carry the same marginal spread. When runoff errors at A and B move together across ensemble members, so they are positively correlated, their contributions compound under routing and the downstream spread is large. When they move in opposite directions, errors cancel and the spread shrinks. The appropriate joint distribution over upstream runoff therefore matters, and the effect is inherently scale-dependent: it intensifies with the number of upstream catchments and with mesh resolution, which controls how finely a basin is partitioned into independently sampled units.


The appropriate joint structure is not universal, but depends on which physical processes dominate the uncertainty. Some sources naturally produce negative correlation between neighbouring catchments. For example, if a coarse gridded rainfall product reports an area average, concentrated rainfall on catchment A implies a deficit at B; the input errors point in opposite directions and tend to cancel downstream. Other sources, for example a systematic gauge under-catch, produce positive correlation. Therefore, we expect that the quantile matching approach may perform well in particular catchments, where the uncertainty is dominated by positively correlated factors, and fail noticeably in catchments where the river flow is negatively correlated between gauges.

We investigate this over Japan using two probabilistic runoff-generation models trained at basin scale and evaluated under both direct basin-scale and routed distributed inference. We hypothesise that independent upstream sampling collapses downstream ensemble spread, and that this effect will grow with the number of upstream catchments. We test whether a simple quantile-matching strategy, which assumes that runoff uncertainty is perfectly spatially correlated across the network, can restore it without over-amplifying the spread. Whether quantile matching proves sufficient, and whether or not it adds skill compared to random matching, is the central question of this study.

\section{Methods}

\subsection{Dataset}\label{sec:dataset}

We use the Japanese Dataset for Distributed River Discharge Prediction (JP-DRDP) \citep{hascoet2026jpdrdp}, which represents Japan as a graph of hydrological catchments connected by upstream--downstream river links. Gauge locations are aligned with catchment outlets, allowing direct evaluation of routed discharge at observed sites.
We retain \textbf{793} gauges whose discharge is sufficiently predictable from meteorological inputs and minimally affected by unresolved dam operations.
Dataset construction, inputs, and gauge selection are detailed in Appendices~\ref{app:data}--\ref{app:inputs}.

\subsection{Runoff training}\label{sec:training}
We train runoff-generation LSTMs at the basin scale over Japan. For each gauge, meteorological forcings and static attributes are aggregated over the full upstream basin,
and the model is supervised against discharge at the basin outlet. This direct basin-scale inference also serves as the reference setting in which no explicit routing or joint upstream sampling is required. This training setup keeps runoff generation separate from routing, so that spatial dependence among upstream members is controlled only by the inference-time sampling strategy.

Both models use the same LSTM backbone and differ only in how they represent predictive uncertainty:

\begin{itemize}
    \item \textbf{Non-parametric.} A random noise vector
    $\xi\sim\mathcal{N}(\mathbf 0,\mathbf I)$ is appended to the LSTM input
    \citep{ruparell2026ai, kraft2026modeling}. Drawing different $\xi$ yields
    distinct runoff trajectories; the network learns a free-form,
    flow-dependent output distribution with no parametric assumption, following the noise-conditioned ensemble approach of \citet{ruparell2026ai}.
    \item \textbf{Parametric.} A two-headed LSTM emits a mean runoff and a dynamic
    spread parameter $\sigma$; members are drawn from a log-normal distribution following these predicted parameters,
    $\hat q\cdot\exp(\sigma z)$ with $z\sim\mathcal{N}(0,1)$.
\end{itemize}

Both models are trained with a deterministic pre-training phase followed by a probabilistic CRPS phase.
Training details and split definitions are given in Appendices~\ref{app:training}--\ref{app:splits}.

\subsection{Inference}\label{sec:inference}

For distributed inference, the trained runoff LSTM is applied separately to each catchment on catchment-scale inputs. The generated runoff is converted into river discharge with DiffRoute, using a Hayami routing scheme \citep{hascoet2026differentiable}. The routing parameters are estimated from a deterministic distributed model and held fixed in the probabilistic experiments.

We then vary only the sampling strategy used to assemble upstream runoff ensembles. random matching draws a separate random seed or standard-normal deviate for every catchment. Quantile matching is implemented as quantile matching across catchments. For each catchment, we first generate its own predictive runoff distribution from local forcings and attributes. We then synchronize ensemble member $m$ by selecting the same probability level in every upstream catchment and routing the resulting set of quantile-equivalent runoff samples; downstream discharge members are therefore sums of aligned quantiles rather than sums of independently ranked samples. The same quantile-matching strategy is used for both runoff models: shared noise members for the non-parametric model and shared standard-normal deviates for the parametric model. This simple joint strategy represents the hypothesis of perfectly positively correlated runoff uncertainty. It is not proposed as a final solution; it is a diagnostic for whether the joint structure matters.

\subsection{Experiments}\label{sec:experiments}

For each runoff family, we evaluate three inference configurations:

\begin{itemize}
    \item \textbf{Direct basin-scale (lumped) inference:} This serves as our performance benchmark or upper bound. Because it predicts the outlet directly without explicit routing, it implicitly captures integrated uncertainties and is expected to yield the highest per-gauge skill.
    \item \textbf{Distributed inference with independent (random) upstream sampling:} This serves as our degraded baseline, representing the naive assumption that upstream uncertainties are uncorrelated.
    \item \textbf{Distributed inference with joint (quantile) upstream sampling:} This is our experimental configuration.
\end{itemize}

By comparing these three, we can quantify exactly how much value is added by introducing spatial correlation (quantile matching) compared to the random baseline, and measure how much of the performance gap with the lumped benchmark is successfully closed.

\subsection{Evaluation Metrics}

To evaluate the quality of our ensemble forecasts, we use a suite of metrics. The metrics assess the probabilistic calibration, deterministic accuracy, and spatial consistency of our models. Our primary probabilistic skill score is the Continuous Ranked Probability Score (CRPS), which we normalise by the standard deviation of observed streamflow to allow comparison across vastly different catchments. To assess ensemble reliability, we calculate the empirical coverage of the 90\% confidence interval and the alpha index, which quantifies deviations from an ideally flat rank histogram. As a guardrail to ensure that probabilistic tuning does not degrade general hydrograph timing and volume, we also report the Nash-Sutcliffe Efficiency (NSE) of the ensemble mean. Formal mathematical definitions for these standard metrics are provided in Appendix~\ref{app:metrics}.

\textbf{Changes in ensemble spread.} 

To directly test our hypothesis regarding spatial correlation, we analyse the absolute width of the ensemble spread (the difference between the maximum and minimum ensemble members). We plot the CDF of the ensemble spread for both the quantile-matched and randomly-matched ensembles. We further disaggregate these metrics as a function of river network size, measured by the number of upstream gauges routing into the prediction gauge. This allows us to quantify exactly how rapidly independent sampling degrades to the mean compared to quantile matching as the routing graph grows in complexity.

Finally, we complement this aggregate view with two illustrative case studies at contrasting scales. We show how performance changes across these catchments of different sizes, as the correlation between gauges will change as the distance between them increases, and the cost of assumptions of gauge independence in random matching, or perfect correlation in quantile matching, may become more apparent.

\section{Results} \label{Results}

\subsection{Spatial sampling dictates probabilistic skill, with no effect on deterministic skill}

\begin{figure}
    \centering
    \includegraphics[width=0.9\linewidth]{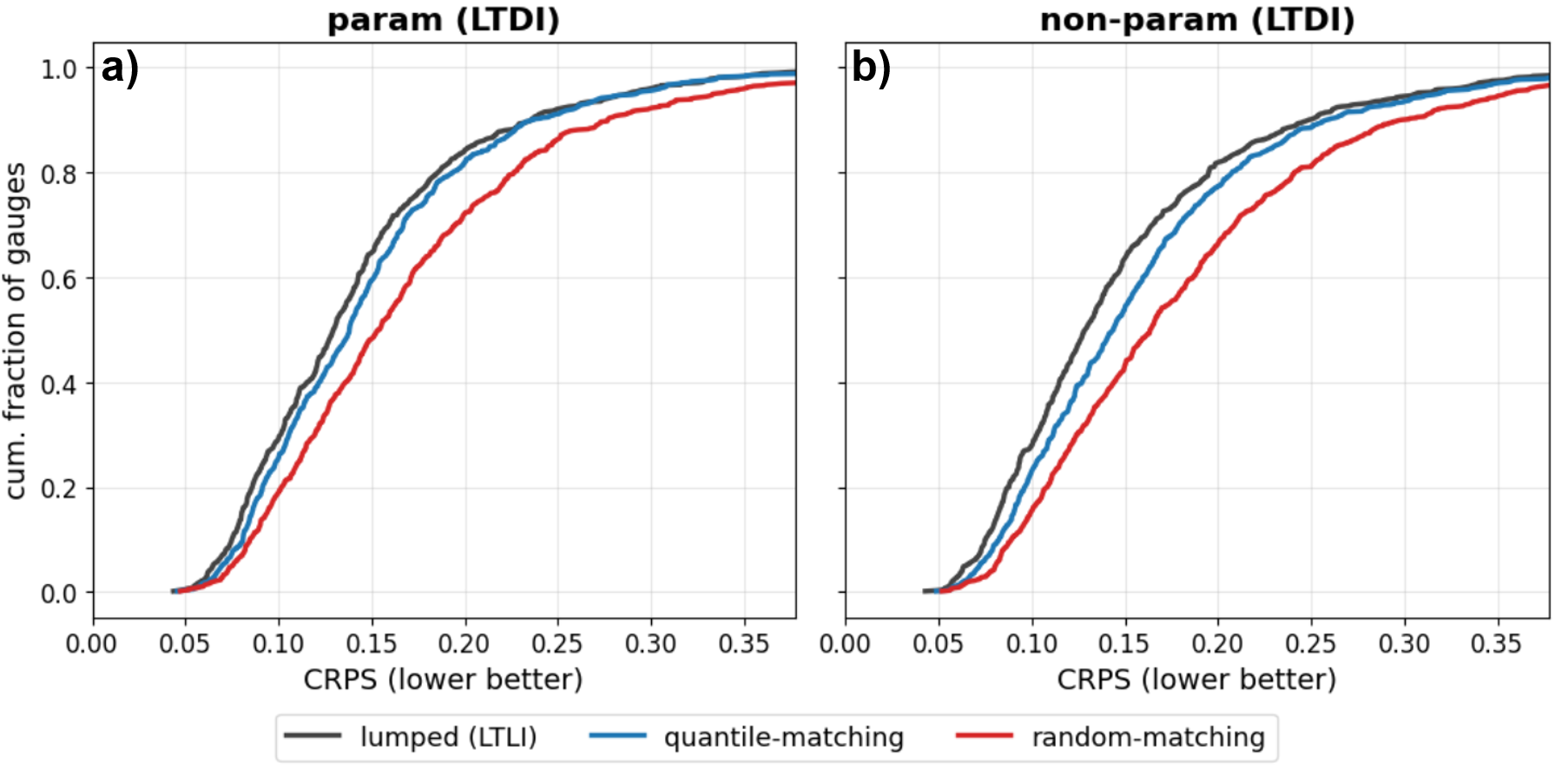}
    \caption{CDF of normalised CRPS scores, for models with parametrically derived ensemble members (left) and non-parametrically derived ensemble members (right). These models generate ensembles in lumped inference (grey), match upstream ensemble members with quantile-matching (blue), or match upstream ensemble members using random-matching}
    \label{fig:CDFs}
\end{figure}

Moving from lumped to distributed routing results in a drop in deterministic skill, as determined by the NSE. However, there is no difference in deterministic performance between the different distributed sampling methods. The ensemble mean NSE remains consistently around 0.84 for all distributed configurations, compared to 0.87 for the lumped models (Tables \ref{tab:metrics_non_parametric} and \ref{tab:metrics_parametric}). The choice of spatial sampling does alter probabilistic skill, however. The Continuous Ranked Probability Score (CRPS) degrades noticeably when moving from lumped models to distributed quantile matching, and degrades even further with random matching. This drop in performance is visible both in the median CRPS values (Tables \ref{tab:metrics_non_parametric} and \ref{tab:metrics_parametric}) and across all percentiles in the cumulative distribution functions of the normalised CRPS scores (Figure \ref{fig:CDFs}). The performance gap between random matching and quantile matching is substantially larger than the gap between quantile matching and the baseline lumped models.

\begin{figure}
    \centering
    \includegraphics[width=0.9\linewidth]{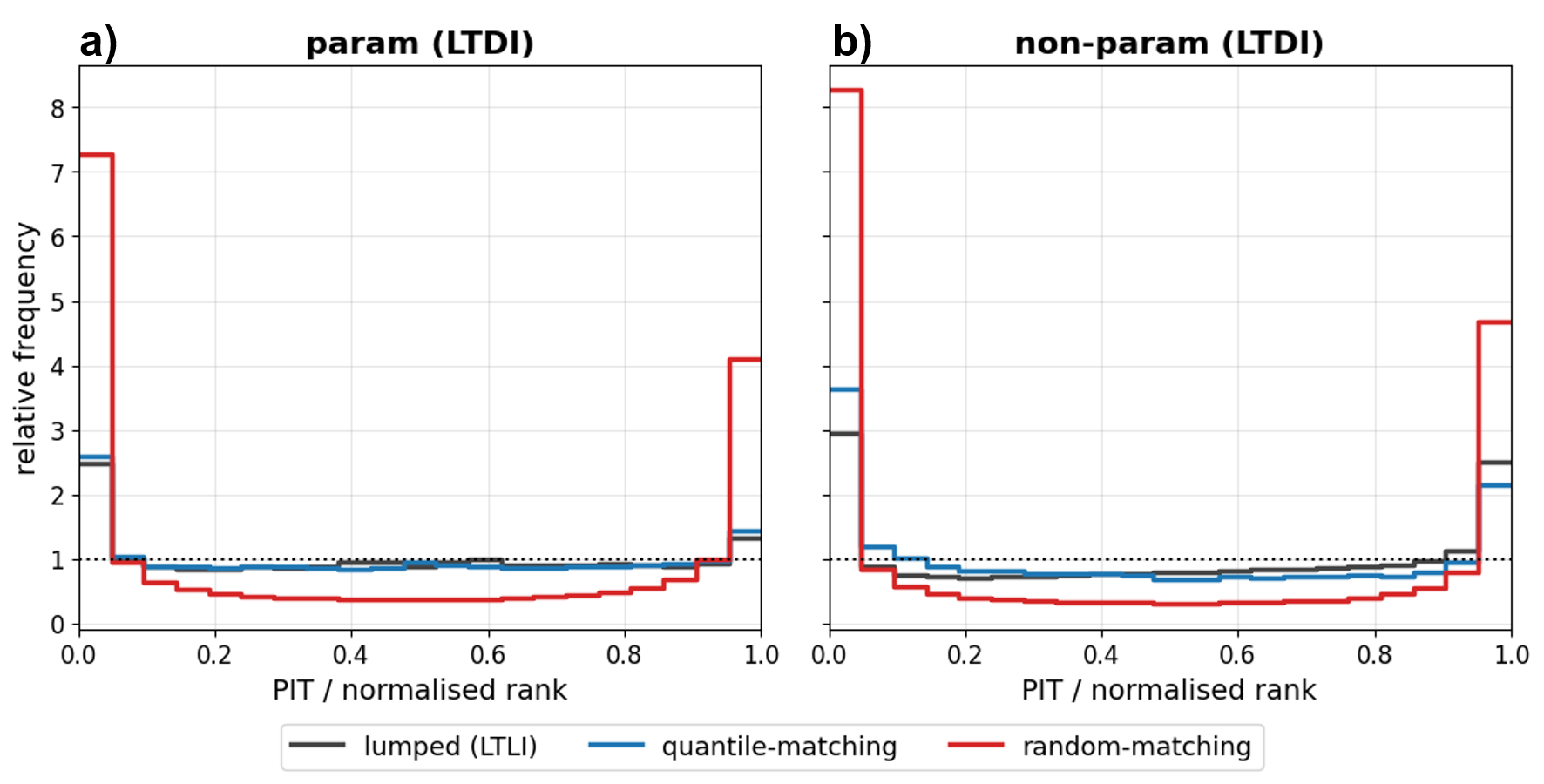}
    \caption{Pooled rank histograms for the seven different modelling approaches, evaluating the reliability and spread of the ensemble forecasts against observed data. The dashed red line in each subplot represents the ideal uniform distribution characteristic of a perfectly calibrated ensemble.}
    \label{fig:rank_histogram}
\end{figure}

This degradation in probabilistic skill under random matching seems to be tied to a collapse in ensemble spread. As seen in the Spread column of Tables \ref{tab:metrics_non_parametric} and \ref{tab:metrics_parametric}, the spread for the random matching models is less than half that of their quantile-matched counterparts. Conversely, the median spread of the quantile-matched and lumped models are the same, and for the parametric models the quantiles too are the same to two significant figures.  This under-dispersion leads to a sharp decline in reliability. The alpha index drops significantly (from 0.72 for quantile matching to 0.44 for random matching), and empirical 90\% coverage falls below 36\%. The pooled rank histograms (Figure \ref{fig:rank_histogram}) confirm this severe under-dispersion under independent sampling, with more than twice as many observations lying entirely outside the ensemble range of the random-matching ensembles compared to the other two methods.

\begin{table}[htbp]
    \centering
    \caption{Median (5\%, 95\%) performance metrics for the non-parametric models. Best median values, across all predictions, for each metric are highlighted in bold.}
    \label{tab:metrics_non_parametric}
    \begin{tabular}{lccccc}
        \toprule
        \textbf{Model} & \textbf{CRPS} & \textbf{NSE} & \textbf{$\alpha$} & \textbf{90\% Cov} & \textbf{Spread} \\
        \midrule
        Lumped     & \textbf{0.13} (0.07, 0.31) & \textbf{0.87} (0.58, 0.95) & \textbf{0.74} (0.34, 0.90) & \textbf{0.68} (0.23, 0.90) & 0.05 (0.00, 0.55) \\
        Quantile & 0.14 (0.07, 0.32)          & 0.84 (0.59, 0.94)          & 0.72 (0.32, 0.90)          & 0.67 (0.21, 0.88)          & 0.05 (0.00, 0.62) \\
        Random & 0.16 (0.08, 0.35)          & 0.84 (0.59, 0.94)          & 0.44 (0.17, 0.78)          & 0.29 (0.06, 0.67)          & 0.02 (0.00, 0.08) \\
        \bottomrule
    \end{tabular}
\end{table}

\vspace{2em} 

\begin{table}[htbp]
    \centering
    \caption{Median (5\%, 95\%) performance metrics for the parametric models. Best median values for each metric are highlighted in bold.}
    \label{tab:metrics_parametric}
    \begin{tabular}{lccccc}
        \toprule
        \textbf{Model} & \textbf{CRPS} & \textbf{NSE} & \textbf{$\alpha$} & \textbf{90\% Cov} & \textbf{Spread} \\
        \midrule
        Lumped     & \textbf{0.13} (0.07, 0.29) & \textbf{0.87} (0.59, 0.95) & \textbf{0.75} (0.38, 0.90) & \textbf{0.78} (0.28, 0.95) & 0.07 (0.01, 0.91) \\
        Quantile & 0.14 (0.07, 0.29)          & 0.84 (0.61, 0.93)          & 0.74 (0.37, 0.90)          & 0.77 (0.29, 0.94)          & 0.07 (0.01, 0.91)\\
        Random  & 0.15 (0.08, 0.33)          & 0.84 (0.61, 0.93)          & 0.51 (0.20, 0.8)          & 0.36 (0.09, 0.73)          & 0.03 (0.01, 0.11) \\
        \bottomrule
    \end{tabular}
\end{table}

\subsection{Patterns are the same regardless of the ensemble generation method}
Tables \ref{tab:metrics_non_parametric} and \ref{tab:metrics_parametric} show that the choice of uncertainty representation does not change the effect of quantile or random matching. This is true across gauges (Figure \ref{fig:CDFs}). While the parametric model shows superior coverage of the observation in the ensemble spread and more uniform rank histograms (Figure \ref{fig:rank_histogram}), quantile matching outperforms random matching in both cases, which in turns is slightly outperformed by the lumped model. Because the degradation in probabilistic skill is tied to the routing and spatial sampling strategies rather than the underlying uncertainty representation, we focus the remainder of our analysis on the parametric models to isolate the effects of quantile matching.

\subsection{Spatial Patterns in Error}
\begin{figure}
    \centering
    \includegraphics[width=0.9\linewidth]{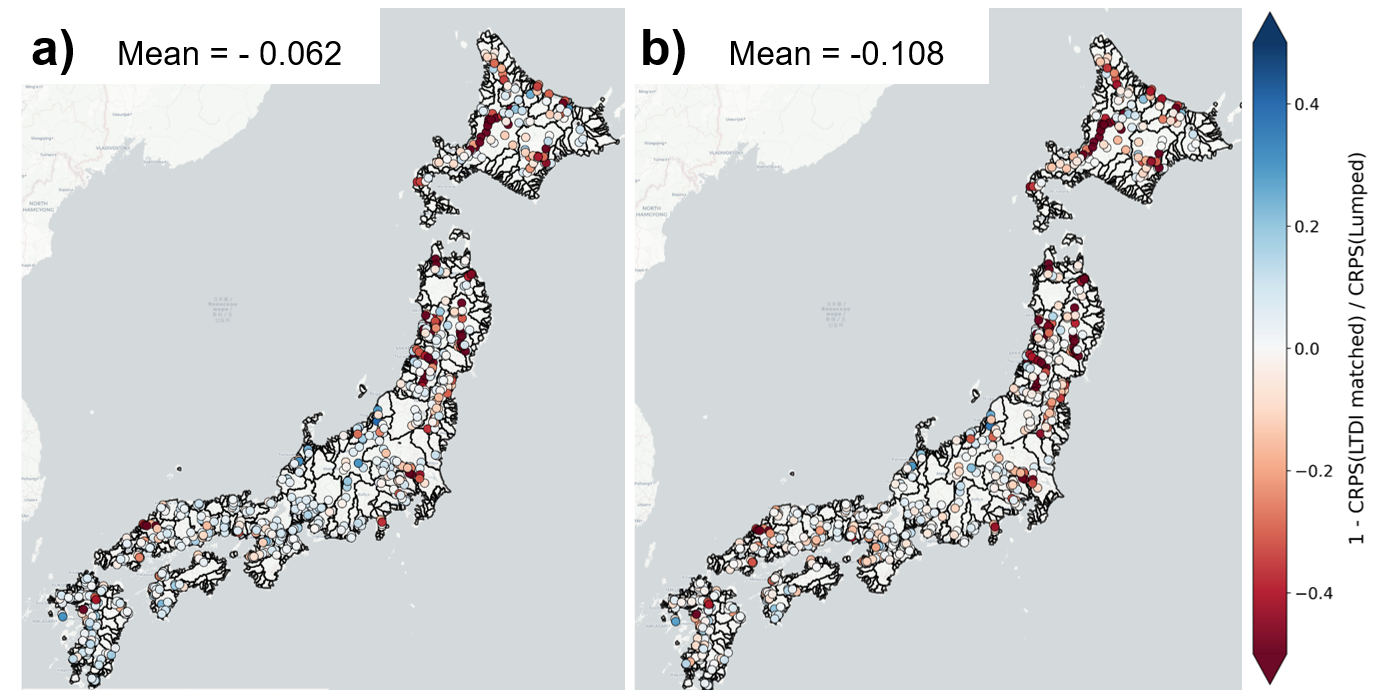}
    \caption{Map showing the normalised CRPSS across all gauges for the parametric models, normalised by the lumped parametric model. On the left we have the model using quantile matching, while the model on the right uses random matching. Gauges where the matching methods outperforms the lumped model are shown in blue, and gauges where the matching underperforms compared to the lumped model are show in red.}
    \label{fig:CRPSS_Map}
\end{figure}

In Figure \ref{fig:CRPSS_Map}, we report the Continuous Ranked Probability Skill Score (CRPSS), treating the parametric lumped model as the reference forecast. Consistent with our initial hypothesis, we observe a general drop in performance when moving from the lumped forecast to a distributed approach. Across our network of 793 gauges, CRPSS is negative in approximately 64\% of gauges when moving from lumped inference to quantile (Figure \ref{fig:CRPSS_Map}a). For random matching (Figure \ref{fig:CRPSS_Map}b), CRPS degrades in over 83\% of gauges. Both methods perform particularly poorly compared to the lumped model in the Hokkaido and Yamagata prefectures, the nort east of Japan. Conversely, the quantile matching model primarily outperforms the random matching approach, and in many cases the lumped model, in western Japan. 

\subsection{Ensemble spread with increasing upstream gauge count}
\begin{figure}[htbp]
    \centering
    \includegraphics[width=\textwidth]{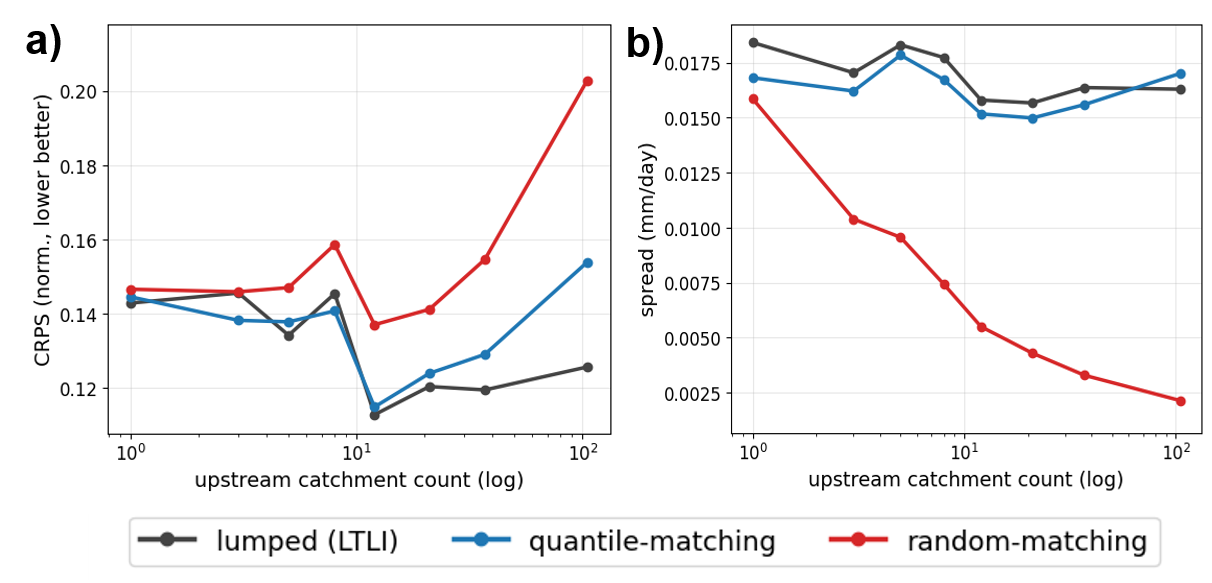}
    \caption{Change in Ensemble Spread (a) and CRPS (b) as the number of upstream gauges increases, for parametrically derived ensembles.}
    \label{fig:gc}
\end{figure}

To examine how this performance gap scales across the network, we analysed model performance as a function of upstream catchment size. Figure \ref{fig:gc} plots the change in CRPS and ensemble spread as the number of upstream gauges increases. As the routing graph grows to include more upstream gauges (from 1 to 356), the quantile-matched ensembles maintain a similar normalised spread (in $mm/day$). The randomly matched ensembles predict a rapidly decreased spread. The disparity between the two methods becomes drastically more pronounced in larger catchments, resulting in a sharp decrease in the ensemble skill, as measured by the CRPS, for the random-matching models as the number of upstream routing steps grows.

\subsection{Demonstration in large catchments}
\begin{figure}
    \centering
    \includegraphics[width=0.9\linewidth]{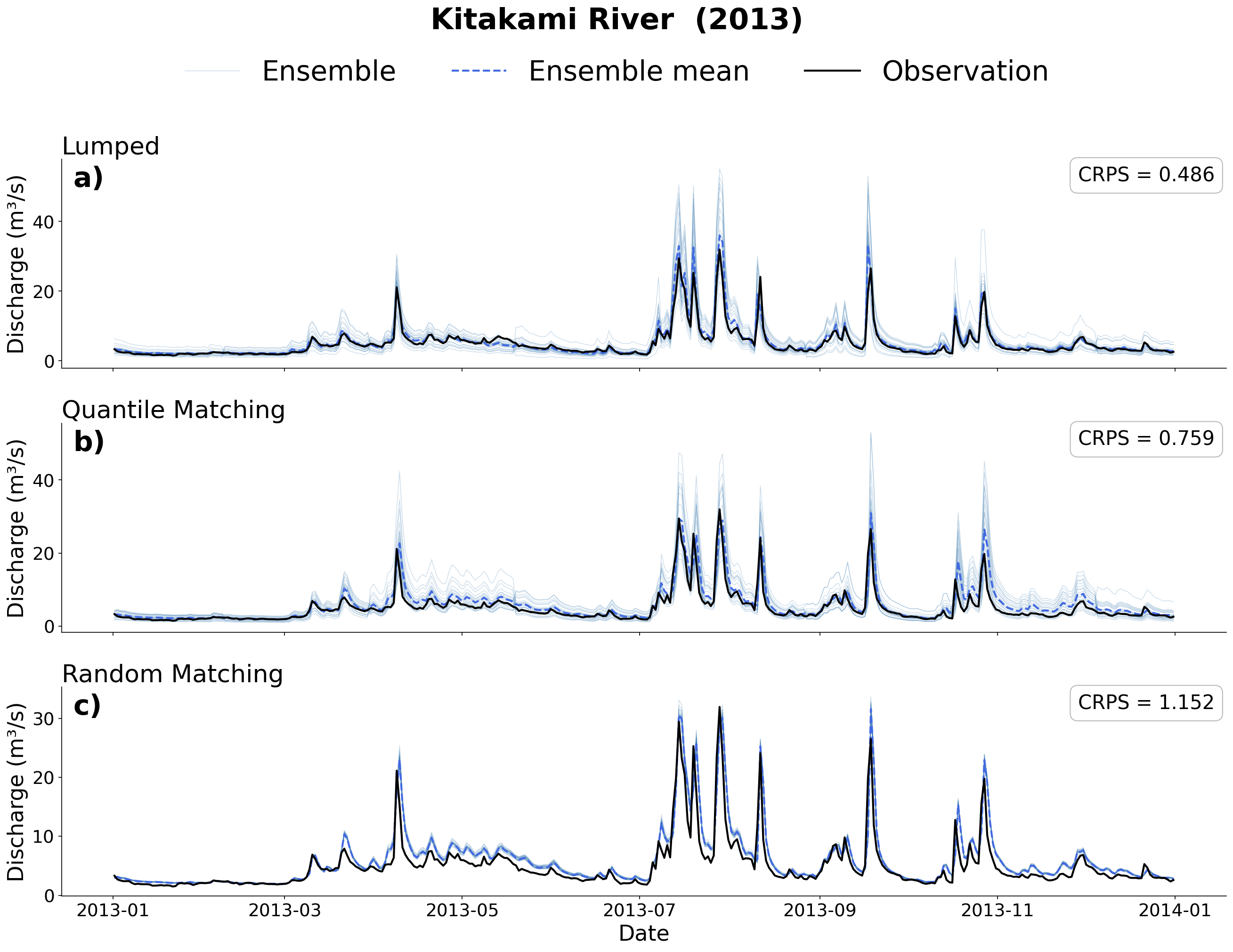}
    \caption{Model performance in Kitakami river. The total upstream area for this gauge is $\approx7700^2$. Hydrographs are as follows: a) Non-parametric model with Joint sampling, b) Parametric model with Joint sampling, c) Non-parametric model with Independent sampling, d) Parametric model with Independent sampling.}
    \label{fig:Big_Hydrograph}
\end{figure}

We can observe the failure of random matching in catchments with many upstream gauges directly in the Kitakami river, with one of the largest upstream areas in our dataset $\approx 7700 km^2$ (Figure \ref{fig:Big_Hydrograph}) and $235$ upstream gauges. All models exhibit a positive bias here, overestimating the true flow. However, because the random matching method is so severely under-dispersed by routing averaging, its entire predicted ensemble falls completely above the observed flow, entirely missing the truth and having the worst CRPS of $1.15$. By preserving spatial correlation, the quantile matched ensembles maintain a much wider range at each timestep, successfully capturing the observations. The CRPS of the quantile matching method is $0.76$, much lower than random matching, but still much larger than the CRPS of the lumped model ($0.49$). From the hydrograph we can see that the quantile matched method is more dispersed than the lumped model for much of the year, particularly during low flow. There is also a positive bias in the quantile matched forecast, with the ensemble mean rising above the observations for most of the period, and more ensemble members overestimating the final two months of flow than for the lumped model.  

\subsection{Limits in small catchments}
\begin{figure}
    \centering
    \includegraphics[width=0.9\linewidth]{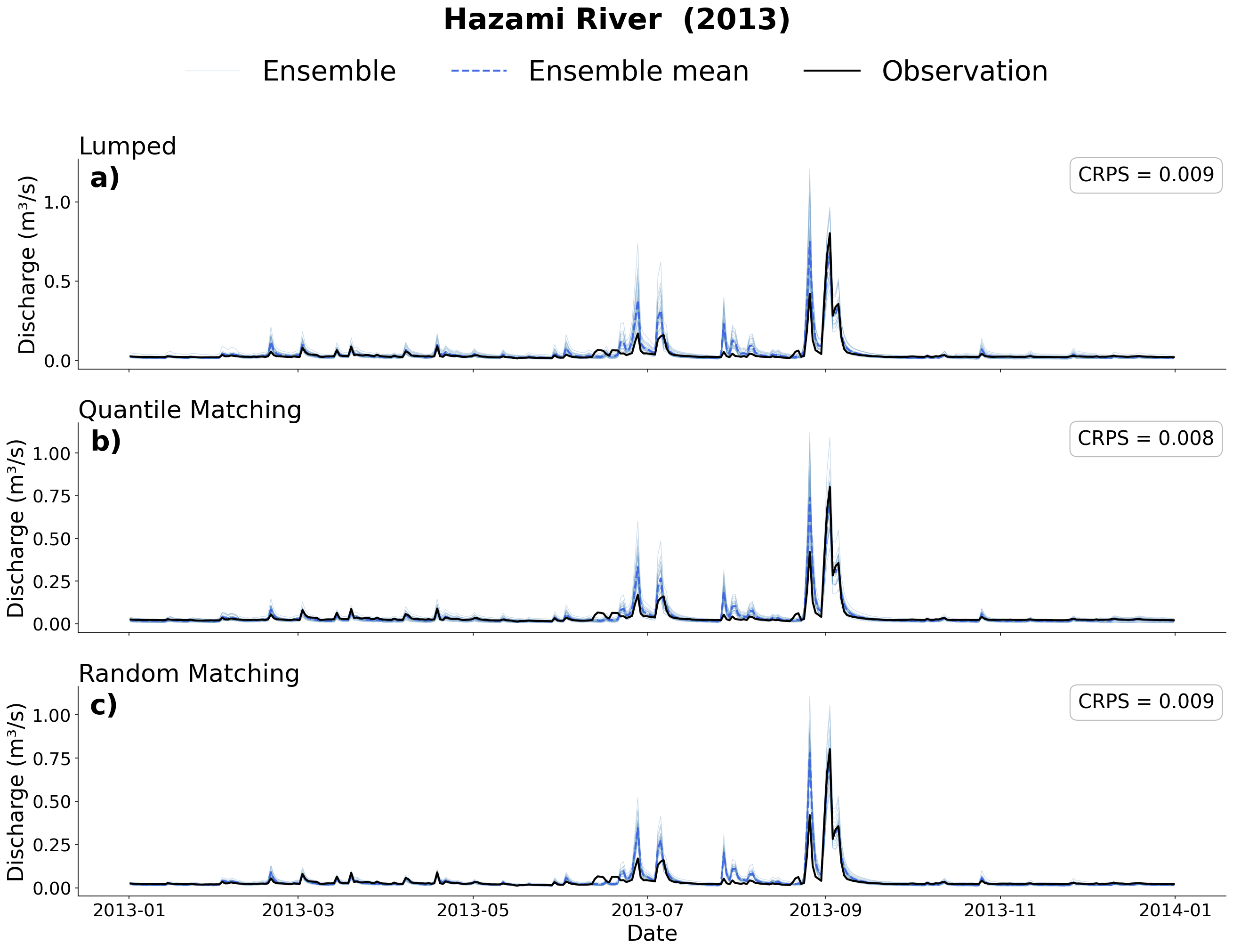}
    \caption{Model performance in Hazama river. The total upstream area for this gauge is $\approx 58km^2$. Hydrographs are as follows: a) Non-parametric model with Joint sampling, b) Parametric model with Joint sampling, c) Non-parametric model with Independent sampling, d) Parametric model with Independent sampling.}
    \label{fig:Small_Hydrograph}
\end{figure}

To examine the limits of this spatial correlation effect, we turn to the Hazama river, a much smaller sub-catchment with an upstream area of roughly 58 km$^2$ (Figure \ref{fig:Small_Hydrograph}) and 4 upstream gauges. Because far fewer upstream gauges route into this point, the penalty for independent sampling is significantly diminished, and the performance gap between the configurations narrows. Here, the three configurations perform almost indistinguishably: the CRPS is 0.009 for the lumped reference, 0.008 for quantile matching, and 0.009 for random matching. Achieving a much lower CRPS overall, the independently sampled models do not suffer the same large collapse in spread seen in larger basins. Despite this general convergence, there remains a slight, persistent advantage to joint sampling over time; from 2010 to 2020, quantile matching outperformed random matching at this gauge in all but two years.

\clearpage

\section{Discussion}

This study set out to test whether the joint spatial structure of upstream ensemble sampling matters for distributed probabilistic streamflow prediction, and whether a simple quantile-matching strategy, which assumes perfect positive correlation across catchments, can recover it where it does. Using Japan as a case study, we compared three inference configurations, direct basin-scale (lumped) inference, distributed inference with independent random upstream sampling, and distributed inference with quantile-matched upstream sampling, across two probabilistic runoff representations. The results confirm that independent sampling collapses downstream ensemble spread through routing-induced averaging, that quantile matching restores most of this spread and, with it, much of the lost probabilistic skill, and that the size of this effect depends strongly on catchment scale and geography. We unpack these findings in turn below: first showing that the effect is invisible to deterministic metrics, then quantifying how much of the gap to the lumped benchmark quantile matching closes, before turning to where the approach succeeds and fails geographically, its scale-dependent limits, and what these results imply for distributed probabilistic hydrology beyond Japan.

\subsection{Deterministic skill is invariant to sampling strategy}

The NSE of the ensemble mean is essentially unchanged across all distributed configurations, holding at approximately $0.84$ for every distributed setup, whether upstream catchments are quantile matched or randomly matched (Tables~\ref{tab:metrics_non_parametric}--\ref{tab:metrics_parametric}). This is expected, as Hayami routing operator is linear map from runoff to discharge, so the expected value of the routed ensemble equals the routed expected value of each catchment's runoff. As long as the sampling strategy used does not change the ensemble mean, the distributed model whose individual runoff models produce well-calibrated means will inherit a well-calibrated distributed mean. However, despite having the same NSE, the random matching approach performs much worse in all probabilistic metrics, such as CRPS, Coverage, and alpha index. Assessing a distributed ensemble model purely on deterministic metrics, like the NSE, can conceal severe probabilistic degradation, such as that demonstrated here by the random-matching configurations. Nuances between distributed probabilistic hydrology models are not seen at the deterministic level, and so analysis must focus primarily on the second and higher moments of the forecast distribution.

\subsection{Quantile matching restores ensemble spread}

The clearest result of this study is that quantile matching essentially restores the ensemble spread of the lumped reference. For the parametric models, median spread under quantile matching matches the lumped median to two significant figures (Table \ref{tab:metrics_parametric}), and the non-parametric models show the same pattern (Table \ref{tab:metrics_non_parametric}). The alpha index also rises from 0.44 to 0.72, close to the lumped value of 0.744 (Table \ref{tab:metrics_non_parametric}). This pattern holds for both uncertainty representations, suggesting that the spread-collapse problem under random matching, and its resolution under quantile matching, is a property of the routing and sampling procedure itself rather than an artefact of how each runoff model represents its predictive distribution. 

This spread recovery translates into a substantial, recovery of probabilistic skill. For the non-parametric model, CRPS improves from 0.164 under random matching to 0.144 under quantile matching, recovering over half of the gap to the lumped reference of 0.129.   In fact, the quantile matching model surpasses the lumped model outright in 36\% of catchments. Figure \ref{fig:gc}a) suggests this benefit is concentrated in catchments on the order of $10$ upstream gauges. We suspect that, as the upstream gauge count falls, the assumption of perfect positive correlation across the contributing catchments becomes a more accurate description of the true correlation structure (nearby catchments share forcing errors more closely than distant ones). In these cases quantile matching can be a better representation of the true joint uncertainty than the lumped model provides. This recovery is not uniform, however, either across catchment scale or across geography, and we examine both of these dependencies in turn below. When there are too few upstream gauges, the impact of the joint sampling is minimal, and all models perform very similarly, as in the case of the Hazama gauge with only 4 upstream gauges.

\subsection{Where quantile matching succeeds and fails geographically}

The spatial pattern of CRPSS (Figure \ref{fig:CRPSS_Map}) supports the correlation-sign hypothesis: quantile matching performs best where runoff uncertainty is positively correlated. Its advantage over random matching, and often even the lumped model, is concentrated in western and southern Japan. Catchments here have fewer upstream gauges (consistent with Figure \ref{fig:gc}) and are closely spaced, making the perfect-correlation assumption highly effective. 

Conversely, both distributed sampling strategies underperform the lumped reference in the Hokkaido and Yamagata prefectures. This shared failure points to a sampling-independent limitation: physical processes that the linear Hayami routing module cannot represent. Both regions feature substantial dam regulation, while Hokkaido is further complicated by wetlands and snowmelt \citep{uno2025hydrological, chapasa2023assessing}. Because lumped models are trained directly against observed outlet discharge, they can implicitly absorb a dam-regulated catchment's characteristic input-output relationship. The distributed models effectively route naturalized runoff, meaning they cannot recover these unrepresented alterations regardless of how the upstream ensembles are sampled. This limitation is likely compounded in Hokkaido by our cross-validation design, which holds out these regions during training, and since there is very little wetland across the rest of Japan is unlikely to be well understood by the model. 

\subsection{Scale-dependence and boundary limits}

The routing-averaging argument predicts that as upstream catchments increase, independent sampling allows progressive cancellation of runoff errors, while perfectly correlated sampling does not. Figure \ref{fig:gc} confirms this: independent models suffer severe spread collapse at scale, whereas joint models appropriately preserve the variance necessary to encompass observed downstream flows.

This scale-dependence is most evident when contrasting large and small basins. In the Kitakami River (Figure \ref{fig:Big_Hydrograph}), which aggregates 235 upstream gauges, random matching's spread collapses entirely. The resulting ensemble range narrowly tracks the ensemble mean, missing the observed flow entirely in the hydrograph showing 2013 discharge. Quantile matching avoids this but overshoots, producing an ensemble that is too wide with a persistent upward winter bias. This occurs because the assumption of perfect correlation begins to break down across such a large, heterogeneous network, and because quantile matching scales up natural snowmelt variance that is, in reality, being stored by upstream dams \cite{KitakamiRiver}. Quantile matching is a clear improvement on random matching's near-total spread collapse, but it does not recover lumped-level skill. The ordering of model performance, lumped outperforming quantile matching which in turns outperforms random matching, is consistent for all years from 2010 to 2020.

At the opposite extreme, the Hazama River has only $4$ upstream gauges. With so little upstream structure, the choice of sampling strategy has a negligible impact on overall spread. However, quantile matching retains a slight edge due to the spatially uniform rainfall of the Baiu front \citep{pc2018hydrological}. Because rainfall across these few gauges is genuinely highly correlated, the joint sampling assumption perfectly aligns with local physical conditions. 

This suggests that quantile matching is most useful in an intermediate regime: where there are enough upstream gauges for joint structure to impact downstream spread, but not so many that spatial heterogeneity invalidates the perfect-correlation assumption. While assuming perfect correlation is a clear improvement over assuming independence, the residual performance gap at large scales like Kitakami demonstrates the limits of this simple approach. This motivates the future development of more expressive models capable of capturing partial or heterogeneous spatial correlations to fully close the gap with lumped models.


\subsection{Generalisation beyond Japan}

The residual gap to the lumped reference at large-scale catchments such as Kitakami shows that even within a setting broadly favourable to quantile matching, perfect correlation is an imperfect approximation once a network grows large and heterogeneous enough. This limitation is likely to be more severe outside Japan, where several of the conditions that make quantile matching a reasonable approximation do not hold. 

The Japanese setting has several characteristics that favour quantile matching. Rivers are short and steep, concentrating runoff quickly and limiting the spatial extent over which independent noise can accumulate. The climate is dominated by large-scale synoptic systems, winter monsoon, typhoon tracks, frontal precipitation bands, whose spatial footprints are large relative to individual catchments, promoting positively correlated forcing errors.  In other settings, these favourable conditions would not hold. In climates dominated by convective precipitation, such as tropical, semi-arid, or mid-latitude summer regimes, the relevant uncertainty is the location and intensity of convective cells whose spatial scale is smaller than the catchment itself. Here, representativeness errors may induce negative correlations between neighbouring sub-catchments.

\section{Conclusion}

Transitioning from lumped to distributed probabilistic hydrology introduces a critical challenge: the explicit handling of spatial uncertainty. When modelling uncertainty sources that lack inherent physical tracking across a landscape—such as observational error, unlike established atmospheric weather ensembles—careful attention must be paid to how ensemble members are matched along a river routing graph. The nearly identical performance between non-parametric and parametric configurations highlights that the choice of marginal distribution modelling at the catchment scale is secondary to the joint distribution modelling across the network. Whether the models draw from an empirical distribution or a parametrically derived one, independent sampling consistently destroys downstream spread, while quantile matching restores it.

As demonstrated in this study, seemingly simple methodological choices, such as assuming perfect spatial correlation (quantile-matching) versus random independent matching, yield drastically different estimates of forecast uncertainty. These severe divergences in probabilistic reliability remain entirely hidden when evaluating only the deterministic ensemble mean. Furthermore, the validity of these sampling assumptions is highly sensitive to catchment scale. While independent sampling catastrophically averages out uncertainty in large routing networks, the rigid assumption of perfect correlation can generate massive over-dispersion in larger basins where gauges have may have very different weather and hydrological conditions.

We believe it is vital for the hydrological machine learning community to move beyond naive spatial sampling strategies. Future developments in distributed probabilistic AI should prioritise dynamic, physically informed spatial correlation structures that can adapt to varying catchment scales. Ensemble copula coupling, widely used in meteorological post-processing \citep{schefzik2013uncertainty}, provides a framework for reassembling marginal distributions with empirically estimated dependence structures without requiring a parametric joint model. Spatially-aware generative models that condition on both local catchment attributes and network position could learn these correlation structures end-to-end from training data. A more immediate practical step would be to estimate spatial correlations empirically from hindcast residuals and use them to construct spatially coherent noise fields for seeding distributed ensembles. Each of these approaches addresses the core challenge identified here: the transition from per-catchment marginal distributions, which existing probabilistic LSTMs already provide, to a joint distribution over the full upstream graph that correctly propagates uncertainty through routing. By explicitly learning and intelligently routing these spatial dependencies we can provide water managers with trustworthy, continuous uncertainty estimates required for operational decision-making across river networks.

\section{acknowledgements}
The authors thank the providers of the observational and meteorological data used in JP-DRDP. Computational resources were provided by RIKEN. We would also like to thank the Advancing Frontier of Earth System Prediction Doctoral Trianing Programme. 

\section{data statement}
The JP-DRDP dataset and the code used in this study will be made available upon
publication. Analysis code for the experiments reported here is archived with the
companion dataset release \citep{hascoet2026jpdrdp}.

\section{Appendix: Data and gauge selection}\label{app:data}
JP-DRDP is built on a point-site-aware segmentation of Japan's four main islands at a $40\,\mathrm{km}^2$ average catchment area, yielding a river-network graph of several thousand non-overlapping catchments whose outlets align with gauge and dam locations \citep{hascoet2026jpdrdp}. River-discharge observations come from the national \textit{in situ} gauge network; we use the quality-controlled set of gauges with calibrated rating curves, aggregated to daily resolution. For the present uncertainty study we further restrict to the \textbf{793} that were shown to be little impacted by dam and artificial waterways.

\section{Appendix: Input specification}\label{app:inputs}
Each catchment carries three categories of inputs, area-aggregated over its polygon
(meteorology) or derived from its geometry (static and channel attributes):
\begin{itemize}
    \item \textbf{Dynamic forcings} (daily): gauge-calibrated radar precipitation
    \citep{nagata2011quantitative} and temperature (with additional meteorological channels as
    available), spatially area-averaged over each catchment polygon. The
    runoff-generation LSTM ingests input sequences of length $T=365$ days.
    \item \textbf{Static catchment attributes}: mean and standard deviation of
    slope, elevation and upstream accumulation area from a high-resolution
    flow-direction map \citep{yamazaki2020high}; physiographic descriptors and a
    fractional land-cover histogram.
\end{itemize}

\section{Appendix: Training schedule}\label{app:training}

All experiments use the same LSTM runoff backbone: two LSTM layers with hidden
size 256, a 365-day initialization window, and daily prediction targets after the
warm-up period. Training is staged. First, deterministic lumped and distributed
mean models are trained from random initialization for 60 epochs, with 40
randomly sampled training windows per epoch. The deterministic phase uses
100-day prediction windows, batch size one, AdamW optimization, runoff learning
rate $10^{-3}$, distributed routing-parameter learning rate $10^{-4}$, and
gradient clipping at norm 1.

The probabilistic models are then initialized from these deterministic means.The non-parametric sampler copies the deterministic LSTM weights and appends a random input channel whose initial weights are zero. The learned-parametric sampler copies the deterministic LSTM and mean head, and initializes its $\sigma$ head from a train-calibrated constant log-normal spread. These probabilistic models are trained for 40 epochs with 20 sampled windows per epoch, minimizing the fair CRPS over $S=8$ training members. The distributed CRPS phase uses 14-day prediction windows to fit in GPU memory.


\section{Appendix: Cross-validation and split definitions}\label{app:splits}

We evaluate spatial generalization with a 10-fold cross-validation over the full set of 793 retained gauges. 
The split is defined at basin level rather than at individual
gauge level: gauges belonging to the same river basin are assigned to the same fold, so that hydrologically connected or closely nested gauges are not divided between training and testing splits.  This prevents information from nearby upstream or downstream gauges in the same basin from leaking into a nominally held-out evaluation site.
For each cross-validation run, one basin-disjoint fold is held out for testing, one separate fold is used for validation, and the remaining eight folds are used for training. Pooling the test predictions across the 10 runs gives one out-of-sample spatial evaluation for every retained gauge. 


\section{Appendix: Evaluation Metric Formulations}\label{app:metrics}

\textbf{Continuous Ranked Probability Score (CRPS)}
The CRPS is widely used in earth sciences to evaluate the skill of probabilistic forecasts \cite{hersbach2020era5}, and has been successfully applied to global generative forecasting models \cite{lang2026aifs}. It generalises the mean absolute error to probabilistic forecasts, rewarding both the reliability (calibration) and sharpness (narrowness) of the predicted distributions. Formally, it is defined as the integrated squared difference between the cumulative distribution function (CDF) of the forecast, $F(x)$, and the empirical CDF of the observation, $y$:

$$ \text{CRPS}(F, y) = \int_{-\infty}^{\infty} (F(x) - H(x - y))^2 dx $$
where $H$ is the Heaviside step function, which equals 1 for $x \geq y$ and 0 otherwise. 
For evaluating an $M$-member empirical ensemble (such as our Seeded models), we compute the unbiased (fair) CRPS estimator:

$$ \text{CRPS}_{fair} = \frac{1}{M} \sum_{i=1}^M |x_i - y| - \frac{1}{2M(M-1)} \sum_{i=1}^M \sum_{j=1}^M |x_i - x_j| $$

where the second term acts as a fair spread correction that prevents the score from being trivially minimised by a degenerate, under-dispersed ensemble. For evaluation, we normalise the per-gauge CRPS by the standard deviation of observed streamflow at that gauge, producing a dimensionless score comparable across catchments of vastly different sizes.

\textbf{Alpha index}
To evaluate the reliability of our ensembles, specifically whether the true observation falls evenly across the predicted ensemble distribution, we use the alpha index, derived from rank histograms (Talagrand diagrams). A perfectly reliable ensemble will have a flat rank histogram, meaning the observation is equally likely to fall between any two ordered ensemble members. The alpha index quantifies deviations from this ideal flatness:

$$    \alpha = 1 - \frac{2}{n} \sum_{t=1}^{n} \left| z^*_t - \frac{t}{n+1} \right|$$

where $n$ is the total number of forecasts, and $z^*_t$ represents the normalised rank of the observation within the sorted ensemble for a given timestep $t$. An alpha index of 1 indicates perfect reliability, while lower values indicate structural biases, such as consistent under-dispersion (U-shaped histograms) or over-dispersion (dome-shaped histograms).

\bibliographystyle{plainnat}
\bibliography{references}

@STRING{AN        = "Astrophys.\ Norv."}

@STRING{MAP       = "Meteor.\ Atmos.\ Phys."}

@article{lang2026aifs,
  title={Aifs-crps: ensemble forecasting using a model trained with a loss function based on the continuous ranked probability score},
  author={Lang, Simon and Alexe, Mihai and Clare, Mariana CA and Roberts, Christopher and Adewoyin, Rilwan and Ben Bouall{\`e}gue, Zied and Chantry, Matthew and Dramsch, Jesper and Dueben, Peter D and Hahner, Sara and others},
  journal={npj Artificial Intelligence},
  volume={2},
  number={1},
  pages={18},
  year={2026},
  publisher={Nature Publishing Group UK London}
}

@article{hersbach2020era5,
  title={The ERA5 global reanalysis},
  author={Hersbach, Hans and Bell, Bill and Berrisford, Paul and Hirahara, Shoji and Hor{\'a}nyi, Andr{\'a}s and Mu{\~n}oz-Sabater, Joaqu{\'\i}n and Nicolas, Julien and Peubey, Carole and Radu, Raluca and Schepers, Dinand and others},
  journal={Quarterly Journal of the Royal Meteorological Society},
  volume={146},
  number={730},
  pages={1999--2049},
  year={2020},
  publisher={Wiley Online Library}
}

@article{hunt2022using,
  title={Using a long short-term memory (LSTM) neural network to boost river streamflow forecasts over the western United States},
  author={Hunt, Kieran MR and Matthews, Gwyneth R and Pappenberger, Florian and Prudhomme, Christel},
  journal={Hydrology and Earth System Sciences},
  volume={26},
  number={21},
  pages={5449--5472},
  year={2022},
  publisher={Copernicus GmbH}
}

@article{ruparell2025hydra,
  title={Hydra-LSTM: A semi-shared Machine Learning architecture for prediction across Watersheds},
  author={Ruparell, Karan and Marks, Robert J and Wood, Andy and Hunt, Kieran MR and Cloke, Hannah L and Prudhomme, Christel and Pappenberger, Florian and Chantry, Matthew},
  journal={Artificial Intelligence for the Earth Systems},
  year={2025},
  publisher={American Meteorological Society}
}

@article{klotz2022uncertainty,
  title={Uncertainty estimation with deep learning for rainfall--runoff modeling},
  author={Klotz, Daniel and Kratzert, Frederik and Gauch, Martin and Keefe Sampson, Alden and Brandstetter, Johannes and Klambauer, G{\"u}nter and Hochreiter, Sepp and Nearing, Grey},
  journal={Hydrology and Earth System Sciences},
  volume={26},
  number={6},
  pages={1673--1693},
  year={2022},
  publisher={Copernicus Publications G{\"o}ttingen, Germany}
}

@article{ruparell2026ai,
  title={AI-generated ensemble river flow forecasting: Using rollout and an additional noise input to build ensemble forecasts},
  author={Ruparell, Karan and Hunt, Kieran MR and Cloke, Hannah and Prudhomme, Christel and Pappenberger, Florian and Chantry, Matthew},
  journal={Authorea Preprints},
  year={2026},
  publisher={Authorea}
}

@article{kraft2026modeling,
  title={Modeling uncertainty with engression: A deep generative time-series approach},
  author={Kraft, Basil and Stalder, Steven and Aeberhard, William H and Ruiz, Nicol{\'a}s Harrington and Meinshausen, Nicolai and Shen, Xinwei and Gudmundsson, Lukas},
  journal={Geophysical Research Letters},
  volume={53},
  number={2},
  pages={e2025GL120122},
  year={2026},
  publisher={Wiley Online Library}
}

@misc{troin2021generating,
  title={Generating ensemble streamflow forecasts: A review of methods and approaches over the past 40 years},
  author={Troin, Magali and Arsenault, Richard and Wood, Andrew W and Brissette, Fran{\c{c}}ois and Martel, Jean-Luc},
  year={2021},
  publisher={Wiley Online Library}
}

@article{schaake2007hepex,
  title={HEPEX: the hydrological ensemble prediction experiment},
  author={Schaake, John C and Hamill, Thomas M and Buizza, Roberto and Clark, Martyn},
  journal={Bulletin of the American Meteorological Society},
  volume={88},
  number={10},
  pages={1541--1548},
  year={2007},
  publisher={American Meteorological Society}
}

@article{cloke2009ensemble,
  title={Ensemble flood forecasting: A review},
  author={Cloke, Hannah L and Pappenberger, Florian},
  journal={Journal of hydrology},
  volume={375},
  number={3-4},
  pages={613--626},
  year={2009},
  publisher={Elsevier}
}

@Article{kratzert2018rainfall,
AUTHOR = {Kratzert, F. and Klotz, D. and Brenner, C. and Schulz, K. and Herrnegger, M.},
TITLE = {Rainfall--runoff modelling using Long Short-Term Memory (LSTM) networks},
JOURNAL = {Hydrology and Earth System Sciences},
VOLUME = {22},
YEAR = {2018},
NUMBER = {11},
PAGES = {6005--6022},
URL = {https://hess.copernicus.org/articles/22/6005/2018/},
DOI = {10.5194/hess-22-6005-2018}
}

@article{kratzert2023caravan,
  title={Caravan-A global community dataset for large-sample hydrology},
  author={Kratzert, Frederik and Nearing, Grey and Addor, Nans and Erickson, Tyler and Gauch, Martin and Gilon, Oren and Gudmundsson, Lukas and Hassidim, Avinatan and Klotz, Daniel and Nevo, Sella and others},
  journal={Scientific Data},
  volume={10},
  number={1},
  pages={61},
  year={2023},
  publisher={Nature Publishing Group UK London}
}

@article{hascoet2026differentiable,
author = {Hascoet, Tristan and Pellet, Victor and Oishi, Satoru and Miyoshi, Takemasa},
title = {Differentiable River Routing for End-to-End Learning of Hydrological Processes},
journal = {Journal of Geophysical Research: Machine Learning and Computation},
volume = {3},
number = {1},
pages = {e2025JH000760},
doi = {https://doi.org/10.1029/2025JH000760},
url = {https://agupubs.onlinelibrary.wiley.com/doi/abs/10.1029/2025JH000760},
eprint = {https://agupubs.onlinelibrary.wiley.com/doi/pdf/10.1029/2025JH000760},
note = {e2025JH000760 2025JH000760},
year = {2026}
}

@article{feng2020enhancing,
  title={Enhancing streamflow forecast and extracting insights using long-short term memory networks with data integration at continental scales},
  author={Feng, Dapeng and Fang, Kuai and Shen, Chaopeng},
  journal={Water Resources Research},
  volume={56},
  number={9},
  pages={e2019WR026793},
  year={2020},
  publisher={Wiley Online Library}
}

@article{bindas2024improving,
  title={Improving river routing using a differentiable Muskingum-Cunge model and physics-informed machine learning},
  author={Bindas, Tadd and Tsai, Wen-Ping and Liu, Jiangtao and Rahmani, Farshid and Feng, Dapeng and Bian, Yuchen and Lawson, Kathryn and Shen, Chaopeng},
  journal={Water Resources Research},
  volume={60},
  number={1},
  pages={e2023WR035337},
  year={2024},
  publisher={Wiley Online Library}
}

@article{schefzik2013uncertainty,
  title={Uncertainty quantification in complex simulation models using ensemble copula coupling},
  author={Schefzik, Roman and Thorarinsdottir, Thordis L and Gneiting, Tilmann},
  year={2013}
}

@article{yang2025global,
  title={Global daily discharge estimation based on grid long short-term memory (LSTM) model and river routing},
  author={Yang, Yuan and Feng, Dapeng and Beck, Hylke E and Hu, Weiming and Abbas, Ather and Sengupta, Agniv and Delle Monache, Luca and Hartman, Robert and Lin, Peirong and Shen, Chaopeng and others},
  journal={Water Resources Research},
  volume={61},
  number={6},
  pages={e2024WR039764},
  year={2025},
  publisher={Wiley Online Library}
}

@article{mosaffa2025gnn,
  title={A GNN Routing Module Is All You Need for LSTM Rainfall--Runoff Models},
  author={Mosaffa, Hamidreza and Pappenberger, Florian and Prudhomme, Christel and Chantry, Matthew and R{\"u}diger, Christoph and Cloke, Hannah},
  journal={EGUsphere},
  volume={2025},
  pages={1--20},
  year={2025},
  publisher={Copernicus Publications G{\"o}ttingen, Germany}
}

@Article{KitakamiRiver, author = {UN.ECAFE},
title =  {A case study of comprehensive development of the Kitakami river basin},
year = 1986,
publisher = {UN.ESCAP},
uri = {https://hdl.handle.net/20.500.12870/8278},
}

@article{uno2025hydrological,
  title={Hydrological Connectivity and Local Environment Alternately Drive Spatial Structure of Floodplain Aquatic Community Across Seasons},
  author={Uno, Hiromi and Utsumi, Shunsuke and Morita, Kentaro and Kishida, Osamu and Alam, Md Khorshed and Negishi, Junjiro},
  journal={Ecology and Evolution},
  volume={15},
  number={2},
  pages={e70880},
  year={2025},
  publisher={Wiley Online Library}
}

@article{pc2018hydrological,
  title={Hydrological simulation of small river basins in northern Kyushu, Japan, during the extreme rainfall event of July 5--6, 2017},
  author={PC, Shakti and Nakatani, Tsuyoshi and Misumi, Ryohei},
  journal={Journal of Disaster Research},
  volume={13},
  number={2},
  pages={396--409},
  year={2018},
  publisher={Fuji Technology Press Ltd.}
}

@article{chapasa2023assessing,
  title={Assessing characteristics and long-term trends in runoff and baseflow index in eastern Japan},
  author={Chapasa, Stanley N and Whitaker, Andrew C},
  journal={Hydrological Research Letters},
  volume={17},
  number={1},
  pages={1--8},
  year={2023},
  publisher={Japan Society of Hydrology and Water Resources}
}

@article{nagata2011quantitative,
  title={Quantitative precipitation estimation and quantitative precipitation forecasting by the Japan Meteorological Agency},
  author={Nagata, Kazuhiko},
  journal={RSMC Tokyo--Typhoon Center Technical Review},
  volume={13},
  pages={37--50},
  year={2011}
}

@article{yamazaki2020high,
  title={High-resolution flow direction map of Japan},
  author={YAMAZAKI, Dai and TOGASHI, Saeka and TAKESHIMA, Akira and SAYAMA, Takahiro},
  journal={Journal of JSCE},
  volume={8},
  number={1},
  pages={234--240},
  year={2020},
  publisher={Japan Society of Civil Engineers}
}

@unpublished{hascoet2026jpdrdp,
  author = {Hascoet, Tristan and others},
  title  = {JP-DRDP: A Japanese Dataset for Distributed River Discharge Prediction},
  note   = {In preparation},
  year   = {2026}
}

\end{document}